\documentclass[conference]{IEEEtran}
\IEEEoverridecommandlockouts

\usepackage{cite}
\usepackage{amsmath,amssymb,amsfonts, mathtools, array}

\usepackage{graphicx}
\usepackage{textcomp}
\usepackage{xcolor}
\usepackage{algorithm}
\usepackage{algpseudocode}
\usepackage{caption}
\usepackage{subcaption}
\usepackage{tabularx}
\usepackage{float}
\usepackage{lipsum,mathtools}
\usepackage{multirow}
\usepackage{dirtytalk}

\usepackage{booktabs}      % \toprule, \midrule, \bottomrule
\usepackage[table]{xcolor} % \rowcolor and \cellcolor
\usepackage{array}         % custom column types

\newcolumntype{L}[1]{>{\raggedright\arraybackslash}p{#1}}
\newcolumntype{C}[1]{>{\centering\arraybackslash}p{#1}}
\newcommand{\thdr}[1]{\textbf{#1}}

\usepackage{url}

\usepackage{amssymb, nccmath}
\usepackage[numbers, sort&compress]{natbib}
\def\BibTeX{{\rm B\kern-.05em{\sc i\kern-.025em b}\kern-.08em
    T\kern-.1667em\lower.7ex\hbox{E}\kern-.125emX}}
    \IEEEaftertitletext{\vspace{-1\baselineskip}}
    \usepackage[left=0.58in, right=0.58in, top=0.71in, bottom=1.05in]{geometry} % safer margins
\begin{document}
%\title{Fuzzy Federated Multi-Label Feature Selection with Reinforcement Learning and Ant Colony Optimization for IoT}
%\title{Fuzzy Federated Multi-Label Feature Selection for IoT: Reinforcement Learning and Ant Colony Optimization}
\title{One Round Is All You Need: Analytic Federated Learning for Task-Heterogeneous Multi-Label Medical Image Classification}
\author{\IEEEauthorblockN{Afsaneh Mahanipour}
\IEEEauthorblockA{\textit{Department of Computer Science} \\
\textit{University of Kentucky}\\
Lexington, KY, USA \\
ama654@uky.edu}
\and
\IEEEauthorblockN{Hana Khamfroush}
\IEEEauthorblockA{\textit{Department of Computer Science} \\
\textit{University of Kentucky}\\
Lexington, KY, USA \\
khamfroush@cs.uky.edu}
}
\maketitle
\begin{abstract}
Federated learning (FL) enables multiple clinical institutions to
collaboratively train a shared disease classifier without centralizing
patient data.
In practice, however, each institution annotates only the pathologies
within its area of expertise, so the federation operates under
\emph{task heterogeneity}: each client holds labels for a strict subset
of the target disease categories while the remaining classes are entirely
unobserved at that site.
Existing gradient-based FL methods fail under this setting because they
require hundreds of communication rounds to converge and because missing
class labels introduce systematic false-negative bias that the model
cannot correct without a principled mechanism.
We propose an analytic federated learning framework
for multi-label medical image classification under task heterogeneity.
The proposed method replaces iterative gradient optimization with three closed-form
operations: a \emph{balanced label projection} that neutralizes
class-imbalance bias by normalizing positive and negative contributions
to equal total mass; a \emph{per-class absolute aggregation law} that
independently assembles the optimal ridge-regression classifier for each
disease category from the sufficient statistics uploaded by its
annotating clients; and an optional \emph{analytic pseudo-label refinement
round} that propagates missing-class knowledge from a confidence-filtered
teacher classifier to non-annotating clients.
The entire procedure requires at most \textbf{two communication rounds},
irrespective of the degree of task heterogeneity or the number of
participating clients.
Experiments on ChestXray14 under four progressively severe
missing-class configurations demonstrate that the proposed method consistently
outperforms the state-of-the-art federated multi-label method FedMLP
by up to \textbf{18.44 BACC points} and \textbf{13.24 AUC points},
while reducing the communication.
Further experiments across different backbone architectures confirm that
the proposed method is backbone-agnostic and generalizes across ResNet, VGG, and
EfficientNet encoders without any hyperparameter adjustment.
\end{abstract}


\section{Introduction}
Automated multi-label disease classification from medical images holds
tremendous clinical value, enabling early and simultaneous detection of
co-occurring conditions from a single radiograph or scan.
However, building accurate classifiers requires large, diverse, and
well-annotated datasets, a resource that individual clinical institutions
rarely possess in isolation.
The straightforward remedy of pooling data across institutions is, in
practice, prohibited by strict patient privacy regulations~\cite{HIPAA},
creating a fundamental tension between the scale required for robust
generalization and the legal constraints on data sharing.

Federated learning (FL)~\cite{FedAvg} has emerged as the principal framework
for resolving this tension.
By keeping raw data at each participating institution (client) and
exchanging only model updates, FL enables collaborative training without
ever centralizing sensitive patient records.
This paradigm has spurred a growing body of work on federated medical image
analysis~\cite{mahanipour2025federated, alkhunaizi2024probing}, demonstrating that
models trained collaboratively across hospitals can approach the
performance of their centralized counterparts on tasks ranging from
organ segmentation to pathology classification.

The overwhelming majority of existing FL methods, however, operate under a
tacit assumption: \emph{task homogeneity}, i.e., every participating client
annotates the same set of disease categories~\cite{FedMLP}.
This assumption is convenient for algorithm design but rarely holds in
clinical reality.
Different hospitals specialize in different conditions: a cardiovascular
center accumulates annotated images of arrhythmias and cardiac
abnormalities, while a pulmonology unit labels pulmonary infiltrates,
pneumonia, and pleural effusions.
When these institutions federate to build a joint classifier, each client
labels only the diseases within its area of expertise, leaving all
other categories \emph{unannotated}, a setting we call
\emph{task heterogeneity}. Sun \emph{et al.}~\cite{FedMLP} formalize this as the federated multi-label
learning problem with partial annotations (FMLL), characterizing each
client's labeled categories as its \emph{active class set} and the unlabeled ones as its \emph{missing class set}.

Under task heterogeneity, na\"ive FL methods fail in two distinct ways.
First, because missing labels are typically treated as negative, the model
at each client is implicitly trained to predict zero for all unlabeled
categories, introducing systematic false-negative bias.
Second, diseases with few annotating clients receive disproportionately sparse supervision
during aggregation.
Existing approaches from federated semi-supervised learning
(FSSL)~\cite{RSCFed,FedFixMatch,FedIRM} and federated noisy-label learning
(FNLL)~\cite{FedNoRo,FedLSR} are not designed for this scenario:
FSSL methods assume that at least some clients or the server hold
complete labels, while FNLL methods model missing entries as symmetric
noise rather than structurally absent annotations~\cite{FedMLP}.
Applying these methods to the FMLL setting therefore results in
suboptimal performance, particularly on rare pathologies.

One of the state-of-the-art methods for this setting, FedMLP~\cite{FedMLP},
addresses task heterogeneity through a carefully designed two-stage gradient
procedure. Although it achieves strong results, its two-stage design has a critical
computational cost: it requires 500 communication rounds.
Moreover, gradient-based training introduces sensitivity to learning rates,
batch sizes, and local optimization steps, none of which have principled
closed-form settings, adding significant hyperparameter burden to a
procedure that already demands careful warm-up tuning.

Analytic learning (AL)~\cite{PIL,RecMP,ACIL} offers an attractive
alternative to gradient-based training.
Rather than iteratively minimizing a loss via backpropagation, AL
reformulates the training of a classification head into a
\emph{least-squares linear regression} problem, for which a closed-form
solution is available in a single epoch. This gradient-free paradigm avoids convergence issues entirely and is
invariant to data ordering and partitioning. However, extending analytic federated learning (AFL) \cite{AFL} to the task-heterogeneous multi-label setting introduces several unresolved challenges: 1) Under raw binary targets, the least-squares solution collapses toward
    always predicting zero for minority classes, since negative samples
    overwhelmingly dominate the squared loss, 2) AFL assumes all label columns are populated for each client.
    Under task heterogeneity, columns for missing classes are absent, not
    zero.
    Treating them as zero actively misleads the regression, introducing
    systematic false-negative bias into the global classifier, and 3) In extreme configurations, e.g., where each class
    has only one annotating client, the single-round AFL has no analogous
    mechanism to propagate knowledge from non-labeling clients.

In this paper, we propose a single-round analytic federated learning framework for multi-label classification under task heterogeneity with partial annotations. The proposed approach provides a closed-form federated solution that addresses the three challenges outlined above while preserving single-round convergence, communication efficiency, and invariance to data partitioning. Our contributions are summarized as follows:

\begin{itemize}
\item We formulate the first closed-form FL framework for multi-label
    medical image classification with partial per-client annotations.
    Our per-class Absolute Aggregation law extends AFL's data-partition
    invariance guarantee to the task-heterogeneous setting.
\item We introduce a principled, closed-form correction for class imbalance
    in analytic multi-label FL.
    Unlike heuristic re-weighting or logit-adjustment schemes, our normalization is directly integrated into the sufficient
    statistics and adds no computational overhead.
\item We propose a two-round analytic procedure in which the Round-1
    classifier serves as a zero-shot teacher for missing-class prediction.
    Confidence-filtered balanced pseudo-label projections are generated
    locally from cached features and aggregated with a single additional
    linear solve, the only gradient-free pseudo-label mechanism in the
    FL literature.
\item The proposed method completes in at most two communication rounds regardless of
    the degree of task heterogeneity.
\item On ChestXray14 dataset under task heterogeneity setting, the proposed method achieves
    competitive and superior performance across all four missing-class
    configurations while
    drastically reducing communication and computational costs.
\end{itemize}

\section{Preliminaries}
\subsection{Federated Learning}
Federated learning (FL) is a distributed machine learning paradigm in which
a set of $K$ participating clients collaboratively train a shared global
model without exchanging their raw data~\cite{FedAvg, mahanipour2025embedded}. Each client $k$ retains a private local dataset
$\mathcal{D}_k = \{({X}_{k,i},\, y_{k,i})\}_{i=1}^{N_k}$, and the
joint training objective is defined over the union
$\mathcal{D} = \bigcup_{k=1}^{K} \mathcal{D}_k$.
By keeping data on-device and sharing only model parameters or statistics,
FL enables collaborative learning in domains where data centralization is
prohibited by privacy regulations, institutional policy, or legal
constraints, circumstances that are ubiquitous in medical imaging~\cite{FedMLP}.

\noindent\textbf{The FedAvg paradigm.}\;
The dominant FL framework, FedAvg~\cite{FedAvg}, operates through an
iterative server–client protocol.
At each communication round $r$, the server distributes the current global
model ${W}^{(r)}$ to all selected clients.
Each client independently runs several steps of stochastic gradient descent
(SGD) on its local objective to produce an updated local model
${W}_k^{(r)}$, which is then uploaded to the server.
The server aggregates the received updates, typically by a weighted
average proportional to local dataset sizes:
\begin{equation}
  {W}^{(r+1)} \;=\; \sum_{k=1}^{K}
  \frac{N_k}{\sum_{j} N_j}\, {W}_k^{(r)},
  \label{eq:fedavg}
\end{equation}
and the process repeats for $R$ rounds until convergence.
Building on this foundation, many methods have been proposed to address
specific limitations: FedProx~\cite{FedProx} adds a proximal regularization
term to bound local update divergence; FedNova~\cite{FedNova} corrects
objective inconsistency arising from heterogeneous local step counts;
and FedDyn~\cite{FedDyn} introduces a dynamic regularizer to align local
and global objectives throughout training.

\subsection{Analytic Learning}
Analytic learning (AL), also known as \emph{pseudoinverse learning}, is a gradient-free training paradigm that derives network weights through
closed-form matrix operations rather than iterative gradient
descent~\cite{PIL,ACIL,RecMP}.
Its origins lie in shallow learning: early networks such as the radial
basis function (RBF) network~\cite{RBF} apply a fixed nonlinear
transformation in the first layer and compute the output weights via a
least-squares (LS) estimate.
Multi-layer extensions of AL generalize this idea,
solving for each linear segment in a deep network analytically by treating
 fixed nonlinear activations of lower layers as a feature
transformation, yielding a one-epoch training procedure without
backpropagation.

In its standard form, AL considers a linearized classification head
$W \in \mathbb R^{d \times C}$ that maps an embedding matrix
$X \in \mathbb R^{N \times d}$ to a label matrix $Y \in \mathbb R^{N \times C}$.
The training objective is a \emph{mean squared error} (MSE) loss:
\begin{equation}
  \mathcal{L}(W) \;=\; \left\|Y - XW\right\|_F^2,
  \label{eq:al_obj}
\end{equation}
whose unique minimum is achieved at the Moore-Penrose (MP)
pseudoinverse solution:
\begin{equation}
  \hat W \;=\; argmin_{W}\;\mathcal{L}(W)
          \;=\; X^{\dagger}Y,
  \label{eq:al_sol}
\end{equation}
where $X^{\dagger} = (X^\top X)^{-1}X^\top$ is the left
pseudoinverse of $X$ (assuming full column rank).
When the full-column-rank assumption is not satisfied, a common occurrence
when the number of samples $N$ is smaller than the embedding dimension $d$, a regularized ridge variant is adopted:
\begin{equation}
  \hat W \;=\; (X^\top X + \gamma I)^{-1}X^\top Y,
  \label{eq:al_ridge}
\end{equation}
where $\gamma > 0$ ensures that the matrix $X^\top X + \gamma I$ is always
positive-definite and invertible.
Eq.~\eqref{eq:al_ridge} is computed in a single pass over the data and
requires no learning rate, no momentum, and no convergence criterion.

\section{Proposed Method}
\subsection{Problem Formulation}
Let \(D=\{D_k\}_{k=1}^K\) denote the complete training corpus distributed across \(K\) clients. Each Clients \(k\) holds a private dataset \(D_k = \{({x}_{k,i},\, y_{k,i})\}_{i=1}^{N_k}\), where \({x}_{i} \in \mathbb{R}^{d_{\text{img}}}\) is an
input image and \(y_{i} \in \{0,1\}^C\) is its multi-hot disease label
vector over \(C\) categories. In realistic clinical deployments, each institution annotates only the
pathologies within its own area of expertise.
We formalise this by defining two disjoint index sets for every client \(k\):
\begin{align}
  AC_k &\;\subset\; [C], \quad \text{(active classes, labeled by client }k) \label{eq:active}\\
  MC_k &\;=\; [C] \setminus AC_k, \quad \text{(missing classes, unlabeled)} \label{eq:missing}
\end{align}
satisfying \(AC_k \cap MC_k = \emptyset\) and \(AC_k \cup MC_k = [C]\).
The server maintains a \emph{class coverage index}
\(S = \{S_c\}_{c=1}^{C}\), where \(S_c = \{k : c \in AC_k\}\) is the
set of clients that label class \(c\), with \(S_c \neq \emptyset\) required
for all \(c\).
Classes with small \(|S_c|\) receive sparse
supervision and represent the principal challenge in the task-heterogeneous setting.

\subsection{Local Phase:}
\noindent\textbf{Feature Extraction:} In this work, all \(K\) clients employ a shared pre-trained backbone network \(f_{backbone}\), parameterized by \(\theta\), to transform their image inputs (e.g., \(x\)) into embedding representations. Unlike gradient-based FL methods that fine-tune the backbone over hundreds of rounds, the proposed method performs a single forward pass per client, eliminating backpropagation entirely. 

For client \(k\), every sample \({x}_{k,i}\) is passed through
\(f_{backbone}\) once to obtain a \(d\)-dimensional feature vector:
\begin{equation}
  h_{k,i} \;=\; f_{backbone}({x}_{k,i},\theta) \;\in\; \mathbb R^{1\times d_e}.
  \label{eq:feat_vec}
\end{equation}

Stacking across all \(N_k\) local samples yields the feature matrix:
\begin{equation}
  H_k \;=\;
  \begin{bmatrix} h_{k,1} \;\cdots\; h_{k,N_k} \end{bmatrix}^\top
  \;\in\; \mathbb R^{N_k \times d_e}.
  \label{eq:feat_mat}
\end{equation}

\noindent where \(d_e\) denotes the embedding length. 

\noindent\textbf{Balanced Label Projection:}
Standard analytic learning formulates the local objective at client \(k\) as a least-squares regression problem, \(\hat{W}_k=H_k^{\dagger}Y_k\), where \({\dagger}\) denotes the Moore-Penrose (MP) inverse and \(Y_k\) is a binary multi-label matrix. A na\"ive formulation directly uses raw binary labels \(y_{k,i}^{(c)} \in \{0,1\}\). However, in medical multi-label classification, severe class imbalance significantly distorts this objective. Rare pathologies (e.g., Pneumonia with approximately 0.9\% prevalence in ChestXray14) are overwhelmingly outnumbered by negative samples. Under the squared loss, the regression is therefore dominated by negatives, causing the learned classifier to bias toward predicting zero for minority classes. This imbalance-driven degeneration resembles the false-negative bias observed in partial-label learning settings.

Additionally, for missing classes $c \in MC_k$, client $k$ possesses no
ground-truth annotation.
Treating missing entries as negative (i.e., $y_{k,i}^{c} = 0$ for
$c \in MC_k$) actively misleads the regressor, as these samples may
genuinely be positive for class $c$, they are simply unknown. 

We resolve both problems simultaneously through a balanced label vector
construction.
For each active class $c \in AC_k$, let
$N_{k,c}^{+} = \sum_{i=1}^{N_k} \mathbf{1}[y_{k,i}^{(c)} = 1]$ and
$N_{k,c}^{-} = N_k - N_{k,c}^{+}$ denote the local positive and negative
sample counts respectively.
The balanced label entry for sample $i$ and class $c$ is:
\begin{equation}
  y_{k,c}[i] \;=\;
  \begin{cases}
    \displaystyle +\frac{1}{N_{k,c}^{+}}
      & \text{if } y_{k,i}^{(c)} = 1 \;\text{and}\; c \in AC_k, \\[8pt]
    \displaystyle -\frac{1}{N_{k,c}^{-}}
      & \text{if } y_{k,i}^{(c)} = 0 \;\text{and}\; c \in AC_k, \\[8pt]
    \displaystyle \;\;0\phantom{+}
      & \text{if } c \in MC_k.
  \end{cases}
  \label{eq:balanced_label}
\end{equation}
The balanced label matrix for client $k$ is then:
\begin{equation}
  Y_k \;=\; \bigl[y_{k,1}\;\cdots\;y_{k,C}\bigr] \;\in\; \mathbb R^{N_k \times C},
  \label{eq:balanced_mat}
\end{equation}
where columns corresponding to missing classes are identically zero.

\noindent\textbf{Local Analytic Computation:}
Given the feature matrix $H_k$ and the balanced
label matrix $Y_k$, client $k$ formulates
its local classification task as a \emph{regularized least-squares} (ridge
regression) problem over its active classes:
\begin{equation}
  \hat{W}_k = \min_{W_k} \;\;
  \left\|{Y_k - H_k W_k}\right\|_F^2 \;+\; \gamma\,\left\|{W_k}\right\|_F^2,
  \label{eq:local_obj}
\end{equation}
where $\left\|{\cdot}\right\|_F$ denotes the Frobenius norm and $\gamma > 0$ is
a ridge regularization coefficient.
The closed-form optimal solution to Eq.~\eqref{eq:local_obj} is the
ridge estimate:
\begin{equation}
  \hat W_k \;=\; \bigl(H_k^\top H_k + \gamma I\bigr)^{-1} H_k^\top Y_k.
  \label{eq:local_sol}
\end{equation}
The regularization term $\gamma I$ serves two roles simultaneously:
(i)~it ensures the matrix $H_k^\top H_k + \gamma I$ is always
positive-definite and invertible, even when $N_k < d$; and
(ii)~it prevents overfitting to the small local dataset available to each client.

Rather than transmitting the full weight matrix $W_k$, which would expose local data statistics, client $k$ computes and sends
only two compact matrices that serve as \emph{sufficient statistics} for the
joint regression problem:

\begin{align}
  A_k &\;=\; H_k^\top H_k \;+\; \gamma\,I,
  \label{eq:Ak} \\[4pt]
  b_{k,c} &\;=\; H_k^\top y_{k,c},
  \quad \forall\, c \in AC_k.
  \label{eq:bk}
\end{align}

$A_k$ is the \emph{regularized autocorrelation matrix} of the local
features, and $b_{k,c}$ is the \emph{balanced label projection vector} for class $c$,
encoding the cross-correlation between the local features and the balanced
label vector.

After completing the local computation, each client $k$
transmits the pair $\{A_k,\;\{b_{k,c}\}_{c \in AC_k}\}$ to the
central server.
The server receives these matrices from all $K$ clients simultaneously.
No model weights, activations, or raw data are transferred.

\subsection{Global Phase: Per-Class Absolute Aggregation under Task Heterogeneity}

In the task homogeneous setting, every client provides annotations for all $C$ classes, and the server can simply aggregate all clients for every class.
Under task heterogeneity, only the clients in $S_c$ provide
non-zero $b_{k,c}$ vectors.
A naive application of the global Absolute Aggregation (AA) law would dilute each class's signal with zero contributions from non-labeling clients. We therefore derive a per-class AA law that aggregates each class independently over only its contributing clients.

For class $c$ with $|S_c|$ contributing clients, the server forms:
\begin{equation}
  A_c \;=\; \sum_{k \in S_c} A_k
              \;-\; \bigl(|S_c| - 1\bigr)\,\gamma\,I
          \;=\; \sum_{k \in S_c} H_k^\top H_k \;+\; \gamma\,I.
  \label{eq:Ac}
\end{equation}
The aggregated label projection for class \(c\) is simply:
\begin{equation}
  b_c \;=\; \sum_{k \in S_c} b_{k,c}
          \;=\; \sum_{k \in S_c} H_k^\top y_{k,c}.
  \label{eq:bc}
\end{equation}
The global classifier weight for class c is then obtained by a single linear system solve:
\begin{equation}
  \hat{w}_c \;=\; A_c^{-1}\,b_c
  \;=\;
  \Bigl(\sum_{k \in S_c}\!H_k^\top H_k + \gamma I\Bigr)^{\!-1}
  \!\!\sum_{k \in S_c}\!H_k^\top y_{k,c}.
  \label{eq:classifier1}
\end{equation}
\noindent and the full classifier matrix is $\hat W = [\hat w_1, \ldots, \hat w_C]$. The aggregation in Eq.~\eqref{eq:classifier1} achieves the following
optimality guarantee, which we state as a formal theorem.

\noindent\textbf{Theorem 1.} Per-Class Aggregation Law under Task Heterogeneity:
Let $H^{[c]} = [H_{k_1};\,\ldots;\,H_{k_{K_c}}]$ be the vertically concatenated feature
matrix of all $K_c = |S_c|$ clients in $S_c$, and let
$y^{[c]}$ be the corresponding concatenated balanced label vector.
Then the proposed aggregation solution in Eq.~\eqref{eq:classifier1}
satisfies: \(  \hat w_c \;=\;
  \bigl(H^{[c]\top} H^{[c]} + \gamma I\bigr)^{-1}
  H^{[c]\top} y^{[c]}\), which is the unique minimizer of the centralized balanced ridge regression
over the joint dataset of all clients in $S_c$.

\begin{figure*}
\includegraphics[width=0.85\textwidth]{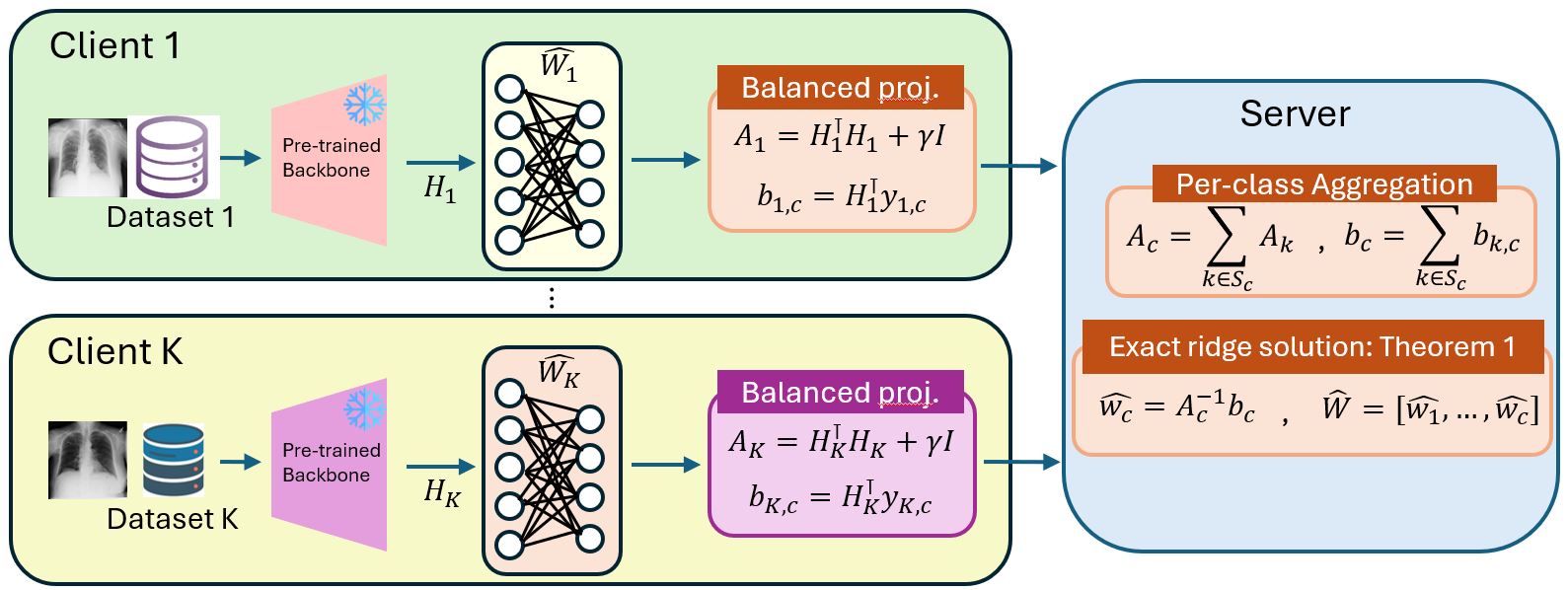}
\centering
\caption{Overview of the proposed framework (Round~1). Each client extracts
features $H_k$ using a shared frozen pre-trained backbone, computes a balanced label
projection to obtain the local sufficient statistics $A_k$ and $b_{k,c}$, and uploads
them to the server. The server performs per-class aggregation over the annotating
client set $\mathcal{S}_c$ and solves a single linear system per class to produce the
globally optimal classifier $\hat{W} = [\widehat{w}_1, \ldots, \widehat{w}_C]$,
as guaranteed by Theorem 1.}
\label{fig1}
\end{figure*}

Unlike gradient-based FL, where client drift under non-IID data degrades
performance and requires many rounds to mitigate, the proposed method achieves the solution in a single communication round regardless of the
data distribution. As depicted in Fig.~\ref{fig1}, each client independently computes and uploads
compact sufficient statistics to the server, which assembles the globally optimal
multi-label classifier $\hat{W}$ in a single closed-form solution without any
gradient exchange.

A limitation of the proposed single-round framework is that the classifier for \emph{cool classes}, those with small $|S_c|$, is trained on very few clients. In contrast, \emph{hot classes} with large $|S_c|$ already benefit from contributions from many clients in the first round and typically achieve strong performance. To close this gap without introducing gradient-based optimization, we propose an analytic pseudo-label refinement procedure that enables an additional round of training. This second round is optional and primarily aims to improve the classifier of underrepresented classes by leveraging pseudo-labeled information from additional clients.

\subsection{Analytic Pseudo-Label Refinement}
After receiving $\hat W^{(1)}$ from the server, each client \(k\) uses the current classifier as a zero-shot predictor to generate pseudo-labels for its \emph{missing} classes $MC_k$. For each missing class $c \in MC_k$, clients \(k\) computes prediction probability of sample $i$ belongs to class $c$ as follows:
\begin{equation}
  p_{k,c}[i] \;=\; \sigma\!\bigl(H_{k}[i,:] \; . \hat w_c^{(1)}\bigr),
  \label{eq:pred}
\end{equation}
where \(\sigma(\cdot)\) denotes the sigmoid function. A sample $i$ is accepted as a pseudo-label only if its prediction is sufficiently confident and far from the decision boundary:

\begin{equation}
  y_{k,c}^{\text{pseudo}}[i] \;=\;
  \begin{cases}
    \displaystyle +\frac{1}{N_{k,c}^{\text{pp}}}
      & \text{if } p_{k,c}[i] > \tau \;\;(\text{pseudo-positive}),  \\[8pt]
    \displaystyle -\frac{1}{N_{k,c}^{\text{pn}}}
      & \text{if } p_{k,c}[i] < 1 - \tau \;\;(\text{pseudo-negative}), \\[8pt]
    \displaystyle \;\;0\phantom{+}
      & \text{otherwise (uncertain/discarded)},
  \end{cases}
  \label{eq:pseudo_label}
\end{equation}
where $N_{k,c}^{\text{pp}} = |\{i : p_{k,c}[i] > \tau\}|$ and
$N_{k,c}^{\text{pn}} = |\{i : p_{k,c}[i] < 1-\tau\}|$ are the
pseudo-positive and pseudo-negative counts. The balanced normalisation in Eq.~\eqref{eq:pseudo_label} mirrors
Eq.~\eqref{eq:balanced_label} ensuring the pseudo-label
refinement does not exacerbate class imbalance. To prevent clients with negligible confident predictions from contributing noisy pseudo-labels, we apply a minimum-count filter.  Client $k$ contributes a pseudo-label projection for class $c$ only if: 
\begin{equation}
  N_{k,c}^{\text{pp}} \;\geq\; \delta_+ \quad \text{and} \quad
  N_{k,c}^{\text{pn}} \;\geq\; \delta_-,
  \label{eq:mincount}
\end{equation}

\noindent where we use $\delta_+ = 5$ and $\delta_- = 50$ throughout. Client \(k\) computes the pseudo-label projection vector for each qualifying missing class:
\begin{equation}
  b_{k,c}^{\text{pseudo}} \;=\; H_k^\top\, y_{k,c}^{\text{pseudo}}.
  \label{eq:pseudo_b}
\end{equation}
and transmits it to the server. The upload volume is at most $|MC_k| \cdot d$ floats per client. 

\noindent\textbf{Re-Aggregation:} The server holds, for each class $c$:
(i) the real-label projections $\{b_{k,c}\}_{k \in S_c}$ uploaded in
Round~1,
(ii) the pseudo-label projections
$b_{k,c}^{\text{pseudo}}$ uploaded in Round~2,
where $P_c^{(2)} \subseteq [K] \setminus S_c$ is the set of clients
that provided valid pseudo-label projections for class $c$, and
(iii) all autocorrelation matrices $\{A_k\}_{k=1}^{K}$ (cached from Round~1).

The server incorporates the feature statistics of \emph{all}
$K$ clients, not just the $|S_c|$ real-label clients.
This is justified because every client contributes a valid $A_k$ encoding
its local feature geometry, regardless of whether it labels class $c$:
\begin{equation}
  A_c^{(2)} \;=\; \sum_{k=1}^{K} A_k \;-\; (K-1)\,.\gamma\,I.
  \label{eq:Ac2}
\end{equation}

The server combines real and pseudo label projections for class $c$:
\begin{equation}
  b_c^{(2)} \;=\;
  \underbrace{\sum_{k \in S_c}\! b_{k,c}}_{\text{real labels (Round~1)}}
  \;+\;
  \alpha\!\underbrace{\sum_{k \in P_c^{(2)}}\! b_{k,c}^{\text{pseudo}}}_{\text{pseudo labels (Round~2)}},
  \label{eq:bc2}
\end{equation}
where $\alpha \in (0, 1]$ is the \emph{pseudo-label weight} controlling
the trust assigned to pseudo-label contributions relative to real ones
(default $\alpha = 0.5$).
When $\alpha = 0$, Eq.~\eqref{eq:bc2} reduces to the Round-1 solution;
when $\alpha = 1$, pseudo-labels are treated with full confidence.
The server solves one linear system per class to obtain the refined classifier:
\begin{equation}
  \hat w_c^{(2)} \;=\; \bigl(A_c^{(2)}\bigr)^{-1}\, b_c^{(2)},
  \label{eq:classifier2}
\end{equation}
and $\hat W^{(2)} = [\hat w_1^{(2)}, \ldots, \hat w_C^{(2)}]$.

\section{Experiments}
We conduct comprehensive experiments to validate the proposed method against
state-of-the-art federated learning methods under the task-heterogeneous
multi-label setting.

\subsection{Datasets}
We evaluate the proposed method on the \textbf{ChestXray14} benchmark~\cite{CXR14},
a publicly available chest radiography dataset assembled from the
National Institutes of Health Clinical Centre containing frontal-view
X-ray images from more than 30,000 unique patients.
Each image may carry multiple simultaneous pathology labels drawn from
14 thoracic disease categories, including Atelectasis, Cardiomegaly,
Effusion, Infiltration, Mass, Nodule, Pneumonia, and Pneumothorax,
among others.
Following the protocol established in FedMLP~\cite{FedMLP}, we restrict
attention to the eight categories with the highest incidence of
positive cases, as these yield a more demanding and clinically
meaningful multi-label classification benchmark.
Posterior-anterior (PA) view images are selected from the full dataset
and partitioned into training and test sets at a 7:3 ratio, with the
test set held out entirely throughout all experiments.

\subsection{Partial Label Generation}
To simulate task heterogeneity across distributed clients, we adopt
the partial label generation protocol of FedMLP~\cite{FedMLP} and
instantiate it across four progressively more challenging missing-class
configurations.
With $K = 8$ clients and $C = 8$ target classes, we generate four
experimental conditions by varying the number of classes withheld from
each client:

\begin{itemize}
  \item \textbf{Missing~1:} each client withholds 1 class, leaving 7
    active classes per client.
    Every class is labeled by exactly 7 clients, providing dense
    cross-client supervision, the mildest setting.
  \item \textbf{Missing~3:} each client withholds 3 classes, leaving 5
    active classes per client.
    The class coverage varies: some classes are labeled by fewer clients,
    creating an early divide between hot and cool classes.
  \item \textbf{Missing~5:} each client withholds 5 classes, retaining
    only 3 active classes per client.
    Most classes are now cool, annotated by only 3 of the 8 clients.
  \item \textbf{Missing~7:} each client withholds 7 classes and labels
    exactly 1 class only.
    In this extreme configuration, every class has $|\mathcal{S}_c| = 1$, a single annotating client, and no client holds a single complete
    multi-label example.
    This is the most stringent test of a method's ability to transfer
    knowledge across task boundaries.
\end{itemize}

Labels are removed at random without replacement within each condition,
subject to the constraint that no class is left entirely unobserved across
the federation (i.e., $|\mathcal{S}_c| \geq 1$ for all $c$).
The same label assignment is used across all comparison methods in
each condition to ensure a fair comparison.
Crucially, missing-class annotations are treated as \emph{structurally
absent}, not as noisy zeroes.

\subsection{Comparison Baseline Methods}
We compare the proposed method against eight representative federated learning
baselines drawn from three methodological families.
Table~\ref{tab:baselines} summarizes their key properties.
All gradient-based baselines use the same frozen DenseNet-121 backbone,
the same training/test splits, and the same partial label assignments
as the proposed method, with hyperparameters set following their respective
published recommendations.

\begin{table}[t]
\centering
\caption{Summary of comparison baselines. ``FL family'' classifies each
method by its core mechanism. Rounds gives the number of
communication rounds. Miss.\ labels indicates how missing
classes are treated.}
\label{tab:baselines}

\resizebox{\linewidth}{!}{
\begin{tabular}{L{1.7cm} L{1.4cm} C{1.0cm} L{2.2cm}}
\toprule
\textbf{Method} & \textbf{Family} & \textbf{Rounds} & \textbf{Miss.\ labels} \\
\midrule
FedAvg~\cite{FedAvg} & Base & 50 & Treated as negative \\
RSCFed~\cite{RSCFed} & FSSL & 50 & Semi-supervised \\
FedFixMatch~\cite{FedFixMatch} & FSSL & 50 & Consistency reg. \\
FedIRM~\cite{FedIRM} & FSSL & 50 & Relational matching \\
CBAFed~\cite{CBAFed} & FSSL & 50 & Dynamic threshold \\
FedLSR~\cite{FedLSR} & FNLL & 50 & Self-regularisation \\
FedNoRo~\cite{FedNoRo} & FNLL & 50 & GMM noise detection \\
FedMLP~\cite{FedMLP} & FMLL & 50 & Prototype MLD + CR \\
\midrule
\textbf{Ours} & \textbf{Analytic} & \textbf{1--2} & \textbf{Per-class AA + pseudo} \\
\bottomrule
\end{tabular}}
\end{table}

\textit{FedAvg}~\cite{FedAvg} is the canonical FL baseline.
At each round, clients run one epoch of local SGD on their active class
labels and upload the resulting model weights to the server, which
averages them proportionally to local dataset sizes.
Missing class labels are treated as negative, causing the aggregated model
to predict zero for cool classes as the degree of label absence grows.
FedAvg serves as the foundational reference against which all other
methods measure their improvement.

\noindent\textbf{Federated Semi-Supervised Learning (FSSL) methods.}

\textit{RSCFed}~\cite{RSCFed} addresses non-IID heterogeneity by
constructing a sub-consensus model from a random subset of clients at
each round, and applies local knowledge distillation to ensure that
the sub-consensus model retains knowledge of classes not directly
represented in the selected subset.
In the partial-label setting, unlabeled class columns are treated as
an unsupervised signal, and the distillation objective transfers soft
predictions for missing classes across the federation.

\textit{FedFixMatch}~\cite{FedFixMatch} adapts FixMatch to the federated
setting by generating weakly and strongly augmented views of each image
at every client.
The weakly augmented view through the current global model produces a
pseudo-label for samples above a fixed confidence threshold; these
pseudo-labels then supervise the strongly augmented view via a
consistency loss.
When combined with FedAvg aggregation, this mechanism can propagate
knowledge of missing classes from high-confidence predictions, provided
the global model is already sufficiently accurate.

\textit{FedIRM}~\cite{FedIRM} introduces an inter-client relation
matching strategy designed specifically for semi-supervised medical image
classification.
Clients exchange relational class embeddings rather than full model
weights at each round, enabling each client to align its local feature
space with the global relational structure.
This alignment is particularly helpful for cool classes, whose
local feature clusters may otherwise drift towards the background
class due to missing-label bias.

\textit{CBAFed}~\cite{CBAFed} employs a class-balanced pseudo-labeling
scheme with a dynamic confidence threshold that adapts over the course
of training.
The dynamic threshold allows the method to be conservative during
early rounds when the global model is unreliable and more aggressive
in later rounds as model quality improves.
CBAFed is designed to handle class imbalance explicitly, making it a
relevant comparator for the imbalanced ChestXray14 setting.

\noindent\textbf{Federated Noisy-Label Learning (FNLL) methods.}

The task-heterogeneous setting can be viewed as a structured
noise scenario: from each client's perspective, every positive sample
of a missing class is mislabeled as negative.
FNLL methods are therefore applicable, as they are designed to detect
and correct label noise across the federation.

\textit{FedLSR}~\cite{FedLSR} applies self-regularization during local
training to prevent the model from over-fitting to noisy negative labels.
A regularization term penalizes low-entropy predictions for samples
near the decision boundary, encouraging the model to output uncertain
rather than confidently incorrect predictions for potentially mislabeled
instances.
In our setting, only the local training stage of FedLSR is active,
since all clients in the task-heterogeneous configuration are affected
by structured label noise.

\textit{FedNoRo}~\cite{FedNoRo} detects noisy clients using a
Gaussian Mixture Model (GMM) fitted to each client's per-sample loss
distribution.
Clients whose loss distributions indicate high noise rates are
down-weighted during aggregation.
In the task-heterogeneous setting, every client is structurally noisy
for its missing classes, so we apply only the second stage of FedNoRo
(the noise-robust aggregation component), following the adaptation
used in FedMLP~\cite{FedMLP}.

\noindent\textbf{Federated Multi-Label Learning (FMLL) method.}

\textit{FedMLP}~\cite{FedMLP} is the state-of-the-art method designed
specifically for the task-heterogeneous multi-label FL setting and serves
as our primary benchmark.
It operates in two stages: a warm-up stage trains the
model with a Weighted Partial-Class (WPC) loss combined with logit
adjustment to suppress false-negative bias and correct class imbalance;
and the second stage uses global class prototypes, computed as class-wise feature centroids aggregated across active
clients, together with a self-adaptive threshold mechanism to generate
pseudo-labels for missing classes.
A consistency regularization loss between the global and local models
prevents catastrophic forgetting of cool-class knowledge during fine-tuning.

\vspace{-2 pt}
\subsection{Evaluation Metrics}
In this paper, we report
three complementary metrics that together capture different aspects of
multi-label classification performance under class imbalance.

\noindent\textbf{Balanced Accuracy (BACC).}\;
For each class $c$, the per-class balanced accuracy is defined as the
arithmetic mean of sensitivity (true positive rate) and specificity
(true negative rate):
\begin{equation}
  \text{BACC}_c \;=\; \frac{1}{2}\!\left(
    \frac{\text{TP}_c}{\text{TP}_c + \text{FN}_c}
    \;+\;
    \frac{\text{TN}_c}{\text{TN}_c + \text{FP}_c}
  \right),
  \label{eq:bacc}
\end{equation}
and the reported BACC is the macro-average over all $C$ classes:
$\text{BACC} = \frac{1}{C}\sum_{c=1}^C \text{BACC}_c$.
BACC is the primary metric throughout all experiments.
Unlike standard accuracy, BACC assigns equal weight to the positive
and negative classes regardless of their relative frequency, penalizing
degenerate classifiers that achieve high accuracy by predicting all
negatives.
This is critical in the task-heterogeneous setting, where cool classes
may contain so few positive samples that even a zero-prediction
classifier achieves high standard accuracy.

\vspace{4pt}
\noindent\textbf{Area Under the ROC Curve (AUC).}\;
For each class $c$, the AUC measures the probability that a randomly
chosen positive sample receives a higher predicted score than a randomly
chosen negative sample.
Formally, it is computed as:
\begin{equation}
  \text{AUC}_c \;=\; \int_0^1 \text{TPR}_c\!\left(\text{FPR}_c^{-1}(t)\right) dt,
  \label{eq:auc}
\end{equation}
and we report the macro-average AUC over all $C$ classes.
AUC evaluates the quality of the \emph{ranking} produced by the predicted
probabilities, independent of the chosen decision threshold.
Unlike BACC, AUC is threshold-agnostic, making it a reliable measure of
the underlying discriminative power of the classifier across the full
range of operating points.

\vspace{4pt}
\noindent\textbf{Mean Average Precision (mAP).}\;
Mean average precision summarizes the precision-recall trade-off for
each class and averages across classes:
\begin{equation}
  \text{mAP} \;=\; \frac{1}{C} \sum_{c=1}^C
    \sum_{k=1}^{n_c} \text{P}_c(k)\,\Delta\text{R}_c(k),
  \label{eq:map}
\end{equation}
where the inner sum accumulates precision $\text{P}_c(k)$ at each
positive sample $k$ in rank order, weighted by the change in recall
$\Delta\text{R}_c(k)$.
mAP is particularly sensitive to the ordering of positive samples at the
top of the ranked list, penalizing methods that achieve high recall only
at low precision thresholds.
It is especially informative for rare pathologies (cool classes) where
only a small number of true positives exist in the test set.

\begin{table*}[t]
\centering
\caption{Performance comparison under different missing-label settings.}
\label{tab:main}
\renewcommand{\arraystretch}{1.35}

\resizebox{\linewidth}{!}{
\begin{tabular}{ll|ccc|ccc|ccc|ccc}
\toprule

 &  & \multicolumn{3}{c}{Missing 1 class} 
 & \multicolumn{3}{c}{Missing 3 classes} 
 & \multicolumn{3}{c}{Missing 5 classes}
 & \multicolumn{3}{c}{Missing 7 classes}
 %& \multicolumn{3}{c}{Average} \\
 \\

\cmidrule(lr){3-5}
\cmidrule(lr){6-8}
\cmidrule(lr){9-11}
\cmidrule(lr){12-14}
%\cmidrule(lr){15-17}

Type & Method 
& BACC & AUC & mAP
& BACC & AUC & mAP
& BACC & AUC & mAP
& BACC & AUC & mAP
%& BACC & AUC & mAP \\
\\

\midrule

Base
& FedAvg
& 64.29 & 70.74 & 0.23
& 59.58 & 68.91 & 0.21
& 50.00 & 63.63 & 0.17
& 50.00 & 58.94 & 0.14
%&  &  &  \\
\\

\midrule

FSSL
& RSCFed
& 64.24 & 70.96 & 0.23
& 62.70 & 69.99 & 0.22
& 58.88 & 68.37 & 0.21
& 52.98 & 64.38 & 0.18
%&  &  &  \\
\\

& FedFixMatch
& 64.09 & 70.52 & 0.22
& 63.66 & 69.66 & 0.22
& 61.76 & 68.21 & 0.21
& 58.47 & 64.76 & 0.18
%&  &  &  \\
\\

& FedIRM
& 65.41 & 72.18 & 0.24
& 64.83 & 71.55 & 0.23
& 63.74 & 70.08 & 0.22
& 59.69 & 66.60 & 0.20
%&  &  &  \\
\\

& CBAFed
& 56.90 & 64.91 & 0.19
& 50.42 & 58.60 & 0.16
& 49.94 & 55.37 & 0.14
& 50.01 & 55.72 & 0.14
%&  &  &  \\
\\

\midrule

FNLL
& FedLSR
& 64.61 & 71.85 & 0.24
& 58.37 & 69.79 & 0.23
& 50.55 & 63.48 & 0.18
& 50.00 & 50.86 & 0.11
%&  &  &  \\
\\

& FedNoRo
& 51.48 & 69.85 & 0.22
& 51.01 & 68.14 & 0.21
& 50.00 & 64.74 & 0.18
& 50.00 & 60.65 & 0.15
%&  &  &  \\
\\

\midrule

FMLL
& FedMLP
& 51.78 & 71.32 & 0.24
& 51.73 & 70.37 & 0.23
& 51.58 & 68.37 & 0.21
& 50.06 & 61.68 & 0.16
%&  &  &  \\
\\

\midrule

Analytic
& Ours (2 rounds)
& 69.56 & 77.02 & 0.2645
& 68.66 & 75.91 & 0.2616
& 67.95 & 74.83 & 0.2453
& 65.96 & 72.39 & 0.2332
%&  &  &  \\
\\

& Ours (1 round)
& 69.69 & 77.08 & 0.2649
& 69.09 & 76.32 & 0.2635
& 68.97 & 76.06 & 0.2547
& 68.50 & 74.92 & 0.2498
%&  &  &  \\
\\

\bottomrule
\end{tabular}
}
\end{table*}

\begin{table*}[t]
\centering
\caption{Performance of the proposed method with different backbone architectures under varying missing-label settings.}
\label{tab:backbone}

\renewcommand{\arraystretch}{1.35}

\resizebox{\linewidth}{!}{
\begin{tabular}{l|ccc|ccc|ccc|ccc}
\toprule

 & \multicolumn{3}{c}{Missing 1 class}
 & \multicolumn{3}{c}{Missing 3 classes}
 & \multicolumn{3}{c}{Missing 5 classes}
 & \multicolumn{3}{c}{Missing 7 classes}
 \\

\cmidrule(lr){2-4}
\cmidrule(lr){5-7}
\cmidrule(lr){8-10}
\cmidrule(lr){11-13}

\textbf{Backbone}
& BACC & AUC & mAP
& BACC & AUC & mAP
& BACC & AUC & mAP
& BACC & AUC & mAP
\\

\midrule

ResNet18 & 63.79 & 69.84 & 0.1665 & 62.88 & 68.64 & 0.1614 & 61.65 & 66.61 & 0.1472 & 59.71 & 63.97 & 0.1314 \\
ResNet34 & 63.41 & 68.83 & 0.1524 & 62.78 & 68.23 & 0.1481 & 60.34 & 64.85 & 0.1337 & 58.95 & 62.97 & 0.1187
 \\
ResNet50 & 62.91 & 68.15 & 0.1552 & 61.77 & 66.78 & 0.1490 & 59.68 & 64.00 & 0.1351 & 57.20 & 60.69 & 0.1102 \\
ResNet101 & 61.80 & 67.11 & 0.1529 & 60.63 & 65.62 & 0.1443 & 58.90 & 62.89 & 0.1294 & 55.77 & 58.57 & 0.1014 \\
ResNet152 & 62.00 & 67.06 & 0.1499 & 61.19 & 66.23 & 0.1461 & 59.09 & 62.93 & 0.1271 & 57.21 & 59.86 & 0.1008 \\

\midrule

VGG11 & 59.66 & 63.53 & 0.1254 & 59.12 & 62.91 & 0.1178 & 56.95 & 60.11 & 0.1076 & 53.39 & 55.21 & 0.0856 \\
VGG13 & 60.11 & 64.42 & 0.1265 & 59.32 & 63.05 & 0.1213 & 57.32 & 60.10 & 0.1086 & 53.85 & 55.50 & 0.0861 \\
VGG16 & 58.99 & 62.66 & 0.1183 & 58.35 & 61.79 & 0.1130 & 55.81 & 58.34 & 0.1009 & 53.61 & 55.30 & 0.0838 \\
VGG19 & 58.08 & 61.19 & 0.1070 & 56.81 & 59.92 & 0.1018 & 55.34 & 57.57 & 0.0951 & 52.78 & 53.74 & 0.0780 \\

\midrule

EfficientNet-B0 & 63.29 & 69.01 & 0.1620 & 61.91 & 67.54 & 0.1552 & 60.30 & 64.59 & 0.1373 & 57.24 & 60.68 & 0.1089 \\
EfficientNet-B1 & 63.86 & 70.09 & 0.1806 & 63.15 & 69.66 & 0.1749 &  61.38 & 66.58 & 0.1552 &  59.02 & 63.64 & 0.1280 \\
EfficientNet-B2 & 63.22 & 68.46 & 0.1583 & 62.35 & 67.36 & 0.1531 & 59.89 & 64.37 & 0.1370 & 57.57 & 61.26 & 0.1126 \\
EfficientNet-B3 & 62.63 & 67.69 & 0.1552 & 61.68 & 66.70 & 0.1475 & 60.04 & 63.65 & 0.1296 & 56.75 & 59.49 & 0.1043 \\
EfficientNet-B4 & 62.96 & 68.29 & 0.1605 & 61.95 & 67.16 & 0.1562 & 59.72 & 63.87 & 0.1368 & 58.27 & 61.59 & 0.1136 \\
EfficientNet-B5 & 63.90 & 69.53 & 0.1731 & 62.64 & 68.15 & 0.1689 & 61.45 & 66.21 & 0.1494 & 58.53 & 62.88 & 0.1274 \\
EfficientNet-B6 & 63.86 & 69.67 & 0.1729 & 63.18 & 68.39 & 0.1668 & 61.44 & 65.83 & 0.1523 & 59.41 & 63.06 & 0.1277 \\
EfficientNet-B7 & 63.80 & 69.45 & 0.1734 & 63.09 & 68.57 & 0.1699 & 62.12 & 66.80 & 0.1515 & 59.30 & 62.89 & 0.1324 \\
\bottomrule
\end{tabular}
}
\end{table*}

\subsection{Implementation Details}
All experiments reported in Table II use the DenseNet-121 model from TorchXRayVision~\cite{TXV}, which is pre-trained on the MIMIC-CXR dataset. For Table III, all backbone architectures are pre-trained on ImageNet. In all experiments, the backbone networks remain frozen and their parameters are not updated during training. Table~\ref{tab:hyperparams} summarizes the hyperparameters specific to the proposed method and their corresponding values. For fair comparison, all gradient-based baseline methods are trained for $R=50$ communication rounds, with one local training epoch per client in each round, following the experimental configuration adopted in~\cite{FedMLP}.

\begin{table}[h]
\centering
\caption{Proposed method hyperparameter settings used in all experiments.}
\label{tab:hyperparams}
\resizebox{\linewidth}{!}{%
\begin{tabular}{lll}
\toprule
\thdr{Parameter} & \thdr{Symbol} & \thdr{Value} \\
\midrule
Ridge regularization        & $\gamma$         & $1.0$ \\
Pseudo-label threshold      & $\tau$           & $0.7$ \\
Pseudo-label weight         & $\alpha$         & $0.5$ \\

Min.\ pseudo-positive count & $\delta_+$       & $5$ \\
Min.\ pseudo-negative count & $\delta_-$       & $50$ \\
Number of classes           & $C$              & $8$ \\

Number of clients           & $K$              & $8$ \\
Communication rounds        & $T$              & $1$ or $2$ \\

Inference threshold         & ---              & $0.5$ \\
\bottomrule
\end{tabular}}
\end{table}

\subsection{Results and Analysis}
Table~\ref{tab:main} presents the quantitative comparison of the proposed method against all baselines across the four missing-label configurations.
All experiments in this table use the DenseNet-121 backbone. It shows that our method consistently achieves the best performance across all three metrics and all missing-class settings. The single-round configuration attains BACC values of 69.69\%, 69.09\%, 68.97\%, and 68.50\% for Missing~1, 3, 5, and 7 classes, respectively, surpassing all gradient-based baselines by a substantial margin. Notably, the proposed method completes training in at most two communication rounds, whereas all baselines require 50 rounds of gradient exchange. This highlights the strong efficiency advantage of the proposed analytic framework while simultaneously improving predictive performance.

\noindent\textbf{Comparison with FedMLP.}
FedMLP represents the most relevant prior method designed for this setting. Nevertheless, the proposed method significantly outperforms it. Under the Missing~1 configuration, our method (1 round) improves over FedMLP by $+17.91$ BACC points, $+5.76$ AUC points, and $+0.0249$ mAP. The gap becomes even larger as label scarcity increases. At Missing~7, FedMLP's BACC drops to 50.06\%, which corresponds to chance-level performance, while its AUC falls to 61.68\%, indicating that the model systematically inverts predictions for some classes under this extreme setting. In contrast, our method (1 round) maintains strong performance with 68.50\% BACC and 74.92\% AUC, representing improvements of $+18.44$ BACC points and $+13.24$ AUC points over FedMLP. This collapse of FedMLP can be attributed to its prototype-guided pseudo-labeling mechanism. When only one client provides labeled examples for a class, the prototypes generated during the warm-up stage become unreliable. Consequently, the pseudo-labels produced from these prototypes propagate noise, and the subsequent consistency regularization amplifies the error over the communication rounds.

\vspace{4pt}
\noindent\textbf{Comparison with FSSL methods.}
Among the federated semi-supervised learning (FSSL) baselines, FedIRM achieves the strongest performance across all conditions, obtaining 65.41\% BACC at Missing~1 and 59.69\% at Missing~7. However, the proposed method (1 round) still surpasses FedIRM by $+4.28$ BACC points at Missing~1, and the gap increases to $+8.81$ BACC points at Missing~7. Other FSSL methods degrade more noticeably as label availability decreases. RSCFed and FedFixMatch perform reasonably under mild label absence (Missing~1 and 3) but deteriorate under more severe settings. CBAFed performs poorly across all configurations, with its dynamic pseudo-labeling threshold appearing poorly calibrated for the severe class imbalance present in ChestXray14. As a result, its BACC approaches chance-level performance from the Missing~3 setting onward.

\vspace{4pt}
\noindent\textbf{Comparison with FNLL methods.}
Among federated noisy-label learning (FNLL) baselines, FedLSR shows competitive performance when only one class is missing (64.61\% BACC at Missing~1) but collapses to chance-level performance at Missing~7 (50.00\% BACC), similar to FedAvg. This suggests that its self-regularization strategy, which is designed to address symmetric label noise, struggles with the asymmetric and structured noise introduced by missing-class scenarios. FedNoRo exhibits a different failure mode: its BACC remains below the FedAvg baseline even at Missing~1 (51.48\% vs.\ 64.29\%). This likely arises from its GMM-based noise detection mechanism, which assigns near-zero trust to clients whose labels appear unreliable. In the task-heterogeneous setting considered here, every client lacks labels for several classes, causing the algorithm to mistakenly classify most clients as noisy and leaving the model with insufficient reliable supervision.

\vspace{4pt}
\noindent\textbf{Robustness to increasing label scarcity.}
A clear trend emerges as the number of missing classes increases from 1 to 7: nearly all gradient-based methods degrade substantially. FedAvg, CBAFed, FedLSR, FedNoRo, and FedMLP eventually converge to chance-level performance (BACC $\approx$ 50\%) under the most extreme setting. In contrast, the proposed method demonstrates remarkable robustness. The single-round configuration loses only 1.19 BACC points as the number of missing classes increases from 1 to 7 (69.69\% $\rightarrow$ 68.50\%). This stability stems from the per-class aggregation property established in Theorem~1. Regardless of how many clients annotate a given class, the server aggregates all available information through a closed-form linear solve that yields the exact centralized-optimal classifier for that class. Gradient-based methods, in contrast, rely on iterative optimization across all clients simultaneously and therefore struggle when supervision becomes highly imbalanced across clients.

\vspace{4pt}
\noindent\textbf{Single-round vs.\ two-round proposed method.}
Interestingly, the single-round configuration consistently outperforms the two-round variant across all missing-class settings and evaluation metrics. For example, under Missing~7, the single-round model achieves 68.50\% BACC and 74.92\% AUC, whereas the two-round model obtains 65.96\% BACC and 72.39\% AUC. This result suggests that the analytic pseudo-label refinement stage does not provide additional benefit on this dataset. Two factors likely contribute to this behavior. First, under severe label scarcity (e.g., Missing~7), the Round-1 classifier trained from a single annotating client per class may lack sufficient discriminative power to produce reliable pseudo-labels, introducing noise into the second round. Second, expanding the autocorrelation matrix from per-class to global scope incorporates feature statistics from all clients, including those whose feature distributions for the missing classes may be substantially out-of-distribution relative to the labeling client. This mismatch can further amplify pseudo-label noise. Consequently, we use the single-round configuration as the primary proposed variant for the remaining analyses.

\noindent\textbf{Effect of Backbone Architecture:} 
To assess the generalizability of the proposed method beyond the domain-specific TorchXRayVision backbone, Table~\ref{tab:backbone} reports results using ImageNet-pretrained architectures spanning three families: ResNet~\cite{ResNet}, VGG~\cite{VGG}, and EfficientNet~\cite{EfficientNet}. All backbone weights remain frozen, and the single-round configuration of the proposed method is used throughout.

The results indicate that both backbone architecture and pre-training domain significantly influence performance. All ImageNet-pretrained models achieve lower accuracy than the domain-specific TorchXRayVision DenseNet-121 used in Table~\ref{tab:main}. For example, the best ImageNet backbone (ResNet-18) reaches 63.79\% BACC at Missing~1, compared to 69.69\% for the TXV backbone, with the gap increasing from 5.90 BACC points at Missing~1 to 8.79 points at Missing~7. This highlights the advantage of domain-aligned pre-training: exposure to chest radiographs during pre-training enables the model to capture structural and texture patterns specific to thoracic pathology. In the analytic framework, where only a linear classifier is learned on top of frozen features, the quality and domain relevance of the learned representations directly determine the final performance.

Across ImageNet-pretrained architectures, shallower ResNets generally outperform deeper ones, with ResNet-18 achieving the strongest results within this family. This trend is consistent with observations from frozen-feature regimes, where deeper models often encode highly task-specific semantics optimized for natural-image recognition, while shallower networks retain more transferable mid-level features. EfficientNet variants perform competitively with ResNets and consistently outperform VGG architectures, with EfficientNet-B1 achieving the highest mAP values across several settings, indicating stronger discrimination for minority classes. In contrast, VGG models exhibit the weakest performance and degrade steadily with depth, approaching chance-level accuracy under severe label scarcity. Despite these architectural differences, the proposed method maintains stable performance across all tested backbones without backbone-specific tuning, demonstrating the method’s robustness and plug-and-play compatibility with a wide range of pre-trained encoders.

\section{Conclusion}
We introduced a gradient-free federated learning framework for multi-label medical image classification under task heterogeneity where different clients annotate different disease categories. By leveraging frozen pre-trained backbones, the proposed method reformulates federated training as a sequence of closed-form analytic operations, including a balanced label projection and a per-class aggregation law that reconstructs the centralized ridge-regression solution from distributed sufficient statistics. This formulation eliminates iterative gradient optimization and reduces training to a small number of communication rounds while remaining robust to highly imbalanced and heterogeneous label distributions.

Experiments on ChestXray14 demonstrate that the proposed method consistently outperforms state-of-the-art gradient-based baselines across all missing-class configurations while requiring several orders of magnitude less communication. In particular, it maintains strong performance even in extreme settings where each class is labeled by only one client, whereas competing methods collapse to near-random performance. These results highlight the potential of analytic federated learning as a communication-efficient and stable alternative to gradient-based training for privacy-sensitive multi-label medical applications.
\section*{Acknowledgement}

This work is funded by career grant provided by the National Science Foundation (NSF) under the grant number 2340075.
\small
\bibliographystyle{IEEEtranN}
\bibliography{References}
\end{document}